\documentclass[review]{elsarticle}
\usepackage{graphicx}
\graphicspath{ {./images/} }
\usepackage{amsmath}
\usepackage{amssymb}
\usepackage{indentfirst}
\usepackage{mathrsfs}

\usepackage{textcomp}
\usepackage[center]{caption}
\usepackage{subcaption}
\usepackage{siunitx}
\usepackage{lineno,hyperref}
\modulolinenumbers[5]

\journal{Elsevier \hspace{3mm}}

\begin{document}
	
	\begin{frontmatter}
		
		\title{ Deep convolutional surrogates and degrees of freedom in thermal design}

		\author[mymainaddress]{Hadi Keramati\corref{mycorrespondingauthor}} 
		\cortext[mycorrespondingauthor]{Corresponding author}
		\ead{hkeramati@uwaterloo.ca} 
		\author[mymainaddress]{Feridun Hamdullahpur}

		\address[mymainaddress]{ Department of Mechanical and Mechatronics Engineering, University of Waterloo \\
			200 University Avenue, West Waterloo, Ontario, N2L 3G1, Canada}


\begin{abstract}
	
	We present surrogate models for heat transfer and pressure drop prediction of complex fin geometries generated using composite Bézier curves. Thermal design process includes iterative high fidelity simulation which is complex, computationally expensive, and time-consuming. With the advancement in machine learning algorithms as well as Graphics Processing Units (GPUs), we can utilize the parallel processing architecture of GPUs rather than solely relying on CPUs to accelerate the thermo-fluid simulation. In this study, Convolutional Neural Networks (CNNs) are used to predict results of Computational Fluid Dynamics (CFD) directly from topologies saved as images. The case with a single fin as well as multiple morphable fins are studied. A comparison of Xception network and regular CNN is presented for the case with a single fin design. Results show that high accuracy in prediction is observed for single fin design particularly using Xception network. Increasing design freedom to multiple fins increases the error in prediction. This error, however, remains within three percent for pressure drop and heat transfer estimation which is valuable for design purpose.

\end{abstract}

\begin{keyword}
	\texttt   Geometric deep learning   \sep Geometry processing \sep Heat exchanger \sep Design freedom \sep Surrogate modeling
	
\end{keyword}

\end{frontmatter}

\section{Introduction}

\par
Engineering design process often includes a series of numerical simulation to explore multiple topologies with a wide range of applications in countless industries \cite{hoyer2019neural,sheikholeslami2019nanofluid}. Design of heat transfer devices is more complex than structural or aerodynamic design because of the additional convection-diffusion equation and its pertinent boundary interaction \cite{feppon2021body}. High fidelity simulation used to solve Partial Differential Equations (PDEs) is CPU-intensive, particularly in presence of complex physics \cite{sanchez2018graph}. Since high fidelity simulation is not important during the concept design process, a high performance estimator can accelerate the design process to a great extent \cite{deng2022self,viquerat2020supervised}.
Deep Learning (DL) and neural networks
provide predictive models for a wide range of applications. Physics-Informed Neural Network
(PINN) and feedforward neural networks, for example, showed promising results as
surrogate models to predict results of computer simulation \cite{chen2021improved, kodippili2022data}. Convolutional Neural Network (CNN) is a class of DL which has
gained popularity in recent years with numerous applications in computer vision, radiology,
and most recently in CFD and design community \cite{hoyer2021neural,lee2021deep,zhuang2021learned,allen2022physical,yamashita2018convolutional, chen2021deep}. CNN can recognize features in variable 2D shapes which makes it an appropriate model to predict simulation results of different physical shapes  \cite{hoyer2019neural,viquerat2020supervised}. CNN also benefits from parameter sharing which leads to smaller number of trainable parameters compared to Fully Connected (FC) Layers \cite{rutkowski2021enhancement}. Guo et al. \cite{guo2016convolutional} suggested CNN surrogate model to predict steady state laminar velocity profile generated using Lattice Boltzmann Method (LBM). They used CNN operating on Signed Distance Functions (SDFs) sampled on a 2D grid. 
Several research studies have been carried out to accelerate time-dependent CFD simulation \cite{vinuesa2021potential,brunton2022applying,mendible2022data}. Accelerated CFD models using CNN were introduced for turbulent flow prediction without the presence of boundaries which results in faster simulation even with fine grids \cite{kochkov2021machine}. These models have wide applications in fluid flow prediction particularly in weather forecast that there is no complex boundary condition or complex topology. Google Deepmind reported a model for velocity and pressure prediction of two dimensional fluid flow of a cylinder with 1-2 orders of magnitude faster computation than the finite element solver \cite{pfaff2020learning}. Continuous convolutions are used for Lagrangian 
fluid simulation without the presence of morphing shapes \cite{ummenhofer2019lagrangian}.

U-Net architecture was used as a tool to reconstruct the CFD results of cylinder flow \cite{ma2021physics, abucide2021data}. Regular CNN was also used for pressure prediction in flow around a cylinder \cite{ye2020flow}. CNNs were applied to two dimensional velocity field estimation of blood flow  in artificial lungs using a dataset with the size of 5000 cases \cite{birkenmaier2020convolutional}. Generation of high fidelity velocity field from low fidelity data was studied using GANs  \cite{pourbagian2021super}. A data-driven model based on FC layers for drag prediction of rectangular obstacle with different aspect ratios representing building geometry was reported with as low as 3.17\% error in prediction \cite{sang2021data}. Graph Neural Network has also shown promising results as a fluid dynamics surrogate model in predicting next-step velocity profile for time-dependent flow \cite{chen2021graph,battaglia2018relational,khan2020survey}. The effort in time-dependent accelerated CFD is to minimize error accumulation that occurs during timesteps of the estimation \cite{han2022predicting, kochkov2021machine}. Error accumulation in next-step prediction particularly in turbulent flow causes distortion from ground truth values after few timesteps. Han et al. \cite{han2022predicting} used pivotal nodes to summarize information of graphs acquired from mesh representation into a latent vector and predicted the next-step velocity using FC layers. The information is then decoded through the pivotal nodes. This method reduced error accumulation and computational cost associated with transient physics prediction.

In design optimization process, time-averaged properties are favorable \cite{kavvadias2015optimal, lee2021deep, yoon2020topology}. Viquerat et al. \cite{viquerat2020supervised} used VGG network for drag prediction of the shapes generated using bezier curve at low Reynolds and steady state condition. Zhou and Ooka used FC layers to predict CFD results for fixed cubic geometry for indoor air flow application \cite{zhou2021neural}. They used parameters of the physics as the input of the neural network and velocity and temperature tensors as the output. They reported less than 12\% error for thermal distribution prediction. Neural Network in FC form was also used for drag coefficient prediction of multiple shapes representing cars \cite{jaffar2020prediction}. Inter-vehicular distance was used as the input of the model to predict drag coefficient.

Previous studies focused on prediction of velocity, pressure, and drag coefficient for constant geometries. In some cases, few geometrical parameters such as aspect ratio were considered. Primary focus of this paper is heat transfer and pressure drop prediction from a dataset of shapes generated based on BREP ( Boundary Representation ). BREP is a method for representing geometry using limits such as curves controlled by control points. BREP facilitates boundary condition implementation which is crucial for thermo-fluid structures where nonlinear PDEs should be solved numerically. BREP also gained attention among researchers in structural topology optimization  \cite{zhang2021explicit,guo2014doing}.

In this paper, modern optimized CNN architectures are used for direct heat transfer and pressure drop estimation of morphable shapes in contrast to previous studies performed on constant geometry. This study focuses on heat transfer and pressure drop prediction directly from images without depending on mesh representation. Xception model along with a network optimized CNN are used in this study. A robust model for heat transfer and pressure drop prediction of different geometries can be used for the next generation design process. By applying this model, the time required for domain heat transfer and pressure drop computation for varied morphable topologies can be reduced from several minutes to few seconds.

\section{Methodology}
Here, a summarized information about the domain is provided. The problem is a conjugate heat transfer in a two dimensional space $ D= \overline{\Omega_f} \cup \overline{\Omega_s} \subset \mathbb{R}^2 $ occupied by multiple solid fins, and an incompressible fluid. Fig. \ref{Design_space_marl} shows the domain.

\subsection{High fidelity simulation}

\begin{figure}[h]	
	\centering
	\includegraphics[width=.6\linewidth]{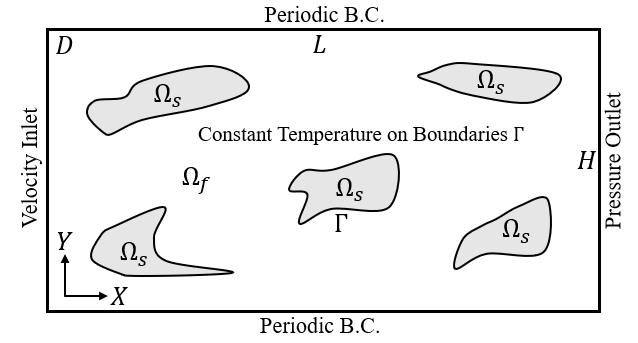}	
	\caption{Setting of computational domain}\label{Design_space_marl}
\end{figure}

A CFD solver is created for the physics shown in Fig. \ref{Design_space_marl}. Incompressible Navier-Stokes equation and  convection-diffusion equation are solved using FEniCS finite element solver. For further details on the numerical method, please refer to Keramati et al \cite{keramati2022deep}. Parallel computing using 16 cores and 32 threads is used for high fidelity simulation in FEniCS.

\begin{figure}[h]
	\centering
	\includegraphics[width=.6\linewidth]{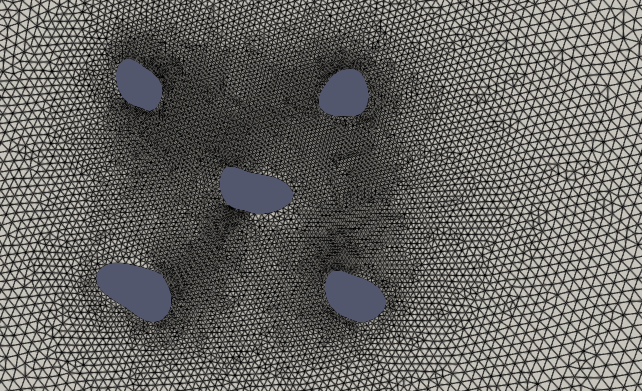}
	\caption{Body-fitted mesh resolution of a selected geometry in multi-shape dataset }	\label{body_mesh_marl}
\end{figure}

\begin{figure}[h]
	
	\centering
	
	\includegraphics[width=0.6\linewidth]{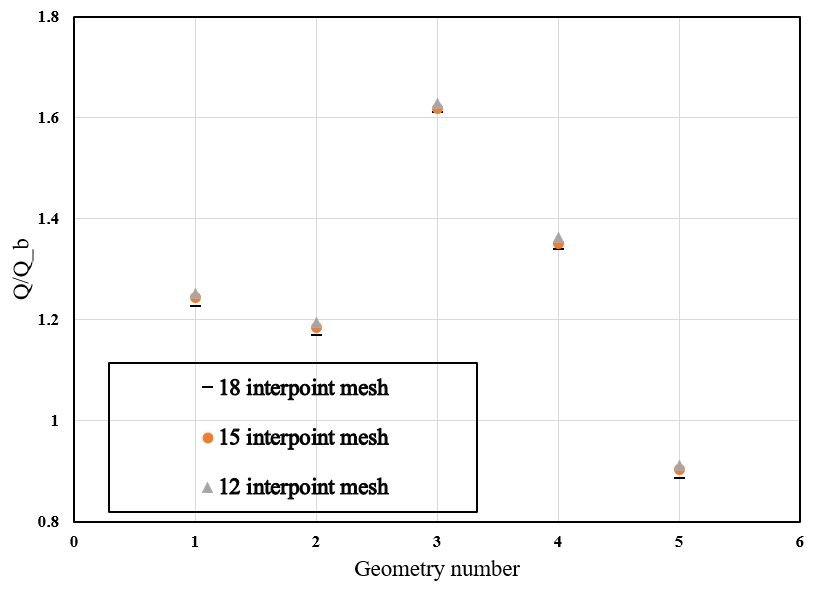}

	\caption{Dimensionless heat transfer for five random geometry settings with different meshing sizes at Re = 10 and Pr = 0.7 }	\label{mesh_heuristics_marl}
\end{figure}

 Here, mesh heuristic for multiple shape inside the domain is presented. Five sets of geometries are considered which are randomly sampled among the shapes constructed during primary running of the framework. A clipping technique is used between number of meshes and mesh size on the curve constructed between points. For the case with single shape, clipping is performed between 15 minimum number of cells and mesh sizes of L/200. However, with the increase in the number of shapes, mesh size is set to be L/150, and number of meshes between points remain higher than 12. Minimum number of meshes for multi-shape framework is decreased compared to the single-agent framework; in case of single shape, the meshes were constructed between the shape with small meshes and boundaries of the domain with larger cells. In case of multiple shapes, meshes are constructed between shapes with small cells on their boundaries which result in high number of meshes. Fig \ref{body_mesh_marl} shows the meshes constructed for a selected geometry. Implementation of this method results in number of grids between 22,000 and 30,000 which depends on the size and shape of the constructed geometry. Figure \ref{mesh_heuristics_marl} shows the mesh heuristics based on the minimum cell numbers between the points. It can be seen that after considering minimum of 12 cells between the points, amounts of heat transfer remain consistent.  

\subsection{CNN models}
Several CNN architectures have been proposed in the past few years \cite{chollet2017xception,he2016deep}. These architectures range from regular CNNs stacked together to develop very deep layers all the way to more complex engineered architecture after google presented inception network \cite{szegedy2015going}. Most of them are competitive in benchmark image classification problems \cite{alzubaidi2021review}. Their performances in real world predictive tasks are, however, apart from the benchmark datasets \cite{fan2022use,brunton2019data}. Xception network and a custom network optimized model are deployed in this study. Xception architecture consists of a modified version of depthwise separable convolution introduced in Inception model to work in a series of operation to improve utilization of computing power \cite{szegedy2016rethinking}. For further details please refer to the original reference \cite{chollet2017xception}.


\subsection{Geometry representation} \label{sec: brep}

BREP is a method for representing shapes by defining topological and geometrical components (e.g. curves, lines and vertices). In this study, we use curves to provide design freedom for fins. Bézier spline is chosen to form continuous fin geometries \cite{chen_design_2017}. Mathematical definition of Bézier spline facilitates the implementation of boundary conditions and conjugate heat transfer \cite{chen_data-driven_2019}. The continuous derivative on the curve also provides a useful tool for Neumann boundary condition implementation \cite{keramati2022deep}. 
\noindent
Bézier curves with a set of control points $ P_i$ with n+1 parameters are defined according to Eq. \ref{bspline}. 
\begin{equation}\label{bspline}
	\alpha(u) = \sum_{i=0}^n P_i B_{i,n}(u) ; u \in [0,1]
\end{equation}
\noindent Where $ B_{i,n}(u)$ is Bernstein polynomials for the \textit{i}'th function of degree \textit{n} which is defined as following:
\begin{equation}\label{bspline2}
	B_{i,n}(u) = {n \choose i} n^i (1-u)^{n-i} ; i=0, .., n
\end{equation}

\subsection{Dataset generation}

\begin{figure}[h]
	\center
	
	\includegraphics[width=0.8\textwidth]{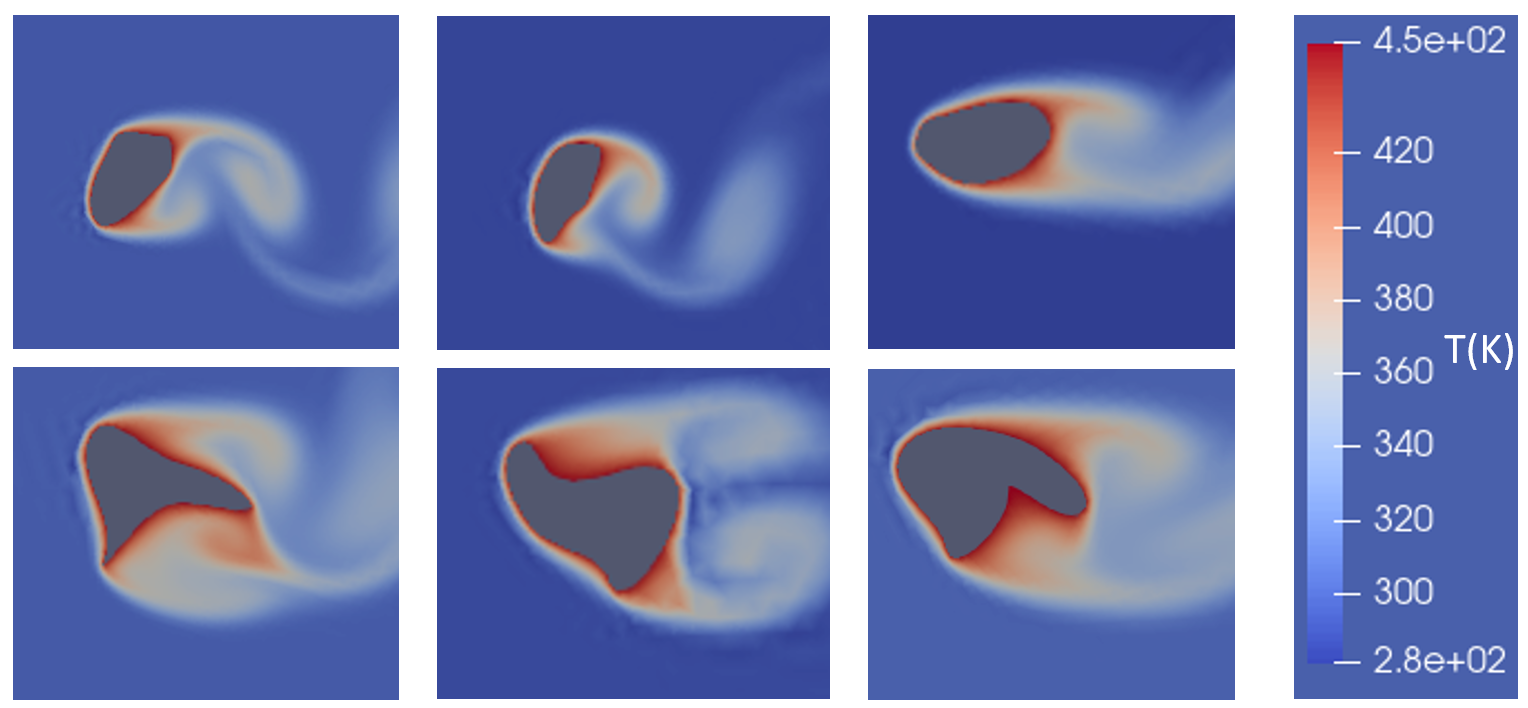}
	\caption{Temperature profile of some random shapes in the single-shape dataset}\label{cnn-data2}
\end{figure}

Geometries are generated using mathematical method explained in section \ref{sec: brep}, which provide 2D space for Eulerian simulation. Each shape
is allowed to occupy a rectangle with the size of H/4 and L/3. Mesh generation is based on adaptive method and heuristics. Multiple scenarios for fluid flow are considered which are controlled by Reynolds and Prandtl number. Since flow is time-dependent, time-averaged numerical result is considered for a long period of time over the third part of the computation. Temperature profile for some of the cases with a single shape at one of the final time instances are shown in Fig. \ref{cnn-data2}. 4000 shapes using four and five points are generated and saved as images. The size and resolution of the images are considered to be constant with a single channel, and 506 $\times$ 506 pixels. CFD results are saved and labeled accordingly. 70\% of the data are used for training and the remainder for cross-validation and test purpose.

\begin{figure}[h]
	
	\centering
	
	\includegraphics[width=.9\linewidth]{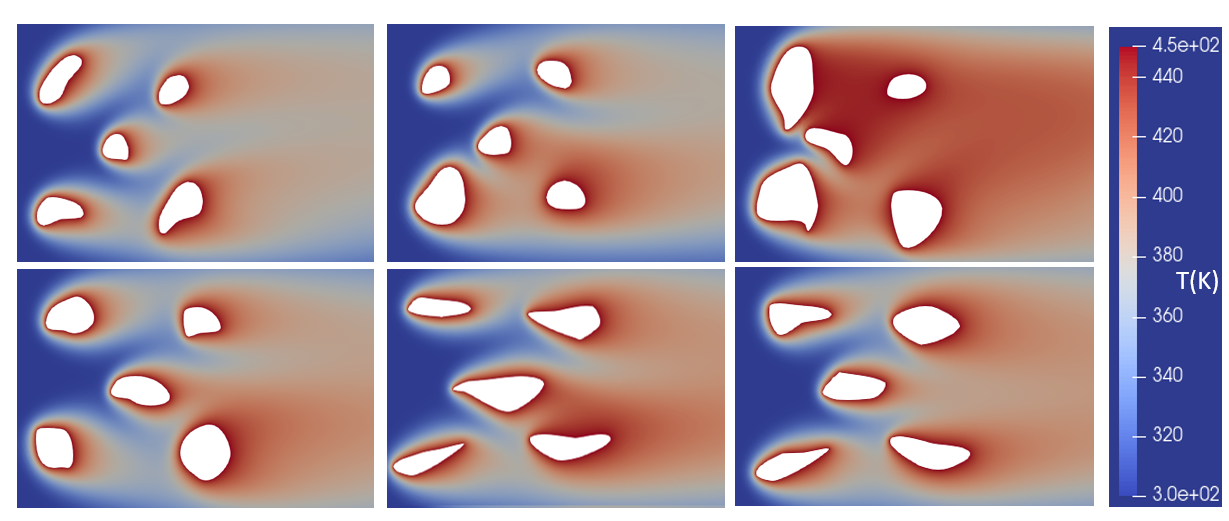}

	\caption{A collection of random shapes in the dataset which is used for training the surrogate model of the physics for the case with multiple shapes inside the domain}	\label{multishape_var}
\end{figure}

34,000 images are generated for multi-shape case using four and five points for each shape. These images along with their CFD results are used for training the surrogate model for multiple shapes in the domain. 90\% of the data are used for training the model and only 10\% for validation and testing the model. Temperature profile for a collection of random geometries from the multishape dataset that is used for training the Xception network is shown in Fig. \ref{multishape_var}. Unity-based normalization is used since heat transfer values for five shapes can go above 1000 watts. 

\subsection{Experimental setup}
All experiments were conducted on a single AMD Ryzen workstation. The workstation was equipped with Ubuntu 20.04 LTS, 16-Core Processor 3.40 GHz CPU, 32.0 GB RAM, and NVIDIA GeForce RTX 3080 Graphics Processing Unit (GPU). Python 3.8.9, CUDA 10.1, TensorFlow, and Keras are used for deep learning implementation.

\section{Results and discussion}

\subsection{Results for a single fin shape}
First, we present the results for the case with a single shape.
\subsubsection{Hyperparameter optimization}

\begin{figure}[h]
	\center
	
	\includegraphics[width=0.99\textwidth]{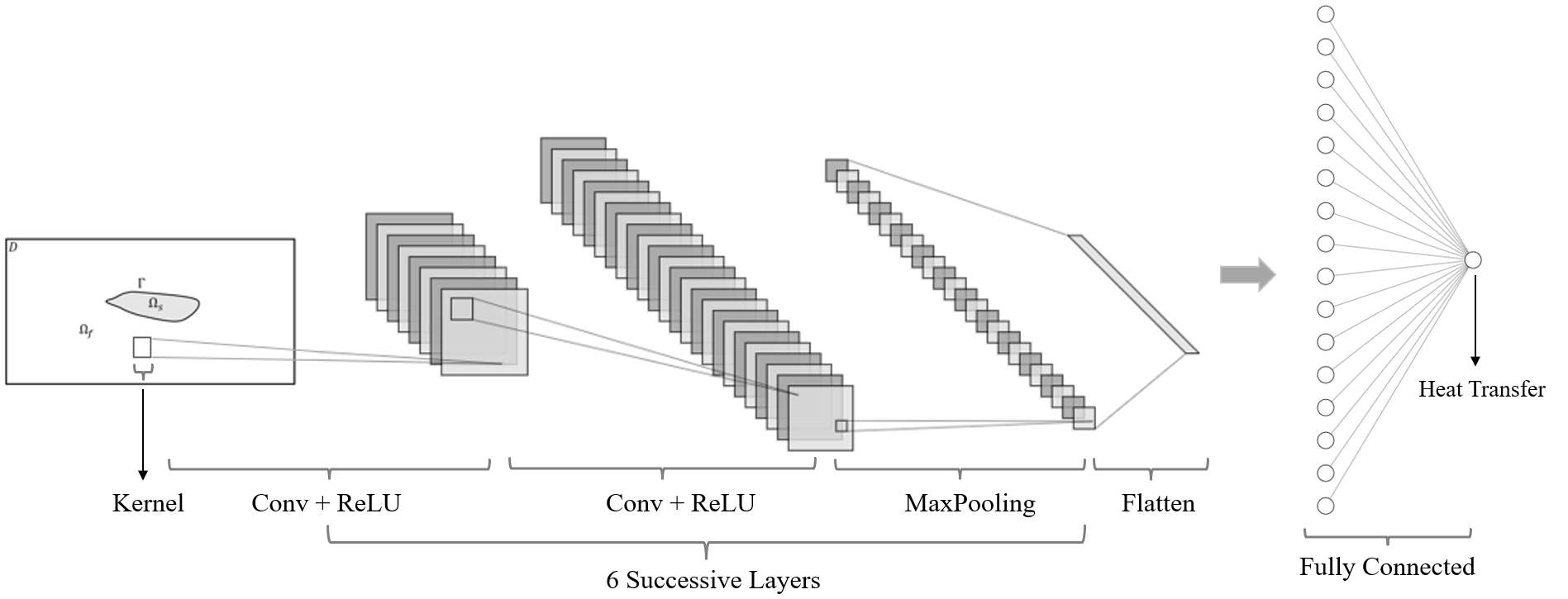}
	\caption{ CNN architecture for heat transfer prediction}\label{cnn2}
\end{figure}

Hyperparameter tuning in DL models is often performed using either grid search or random search \cite{li2018massively}. In this study, random search is used for number of feature extraction layers ($3 < $ feature extraction layers $ <10$), number of filters in each layer ($8 < $ number of filters $ <96$), and number of FC layers ($2 < $ FC layers $ <5$) . After few experiments with kernel size for the convolution operators, it is found that changes in accuracy and convergence between 3 $\times$ 3 and 5 $\times$ 5 kernel size are negligible since the location of the information is in the center of the image. For efficient use of computational resources, 3 $\times$ 3 kernel size for convolutional layers and 2 $\times$ 2 kernel size for maxpooling layers are used.  
The schematics of the CNN architecture used for direct heat transfer prediction is shown in Fig \ref{cnn2}.  It consists of six successive feature extraction layers with two convolutional layers and one max pooling layer in each of them. Some of the hyperparameters of the model are shown in Appendix \ref{Appendix: cnnhyperparameters} . Total number of trainable parameters for the proposed model is 1,160,305. Architecture of the optimized network is shown in Appendix\ref{Appendix:c}. Xception model consists of more than 17 million parameters. Hyperparameter values for Xception model are considered to be the same as those explained in the original article with the exception of those shown in Appendix \ref{Appendix: xception} \cite{chollet2017xception}. Batch size for Xception is considered to be 256 while batch size of 128 leads to the best performance in regular CNN. Training time for heat transfer prediction using regular CNN and Xception network on our experimental setup are 45 minutes and 135 minutes, respectively. The same method and architecture is used for pressure drop considering the output neuron as pressure drop value. Training for pressure drop estimation takes almost one hour for regular CNN and an average of 170 minutes for Xception model. 

\subsubsection{Error behavior}
Root Mean Squared Error (RMSE), Mean Square Error (MSE), and Mean Absolute Error (MAE) are used frequently in the literature for loss function definition to evaluate the quality of a machine learning model during training \cite{han2022predicting}. In this study, MSE is used as given in Eq. \ref{eq: mse} to train the models \cite{han2022predicting}. In this equation, $\hat{{{{y}}}_i}$ is the predicted value, and ${{{{y}}}_i}$ is the ground truth value from high fidelity simulation. $N$ is the number of data points in the batch sample.

\begin{equation}\label{eq: mse}
	MSE({{{{y}}}_i},{{\hat{{y}}}_i}) = \frac{\Sigma_{i=1}^{N} ( {{{{y}}}_i} -{{\hat{{y}}}_i})^2}{N}
\end{equation}

\begin{figure}[h]
	\center
	
	\includegraphics[width=0.5\textwidth]{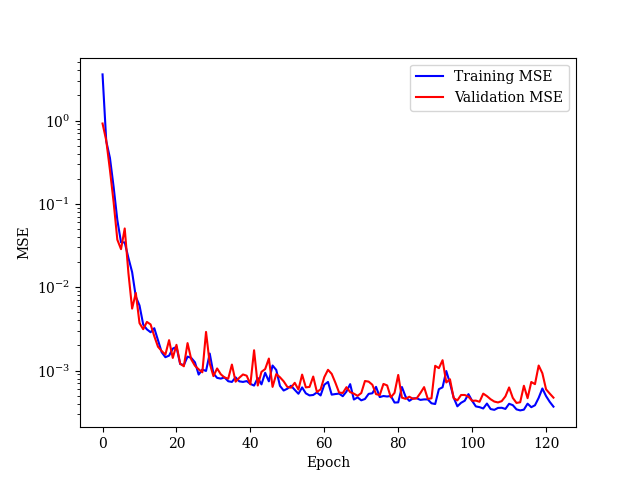}
	\caption{MSE value as a function of epoch during the training of optimized CNN architecture for heat transfer prediction of single shape}\label{cnn-loss2}
\end{figure}

\begin{figure}[h]
	\center
	
	\includegraphics[width=0.5\textwidth]{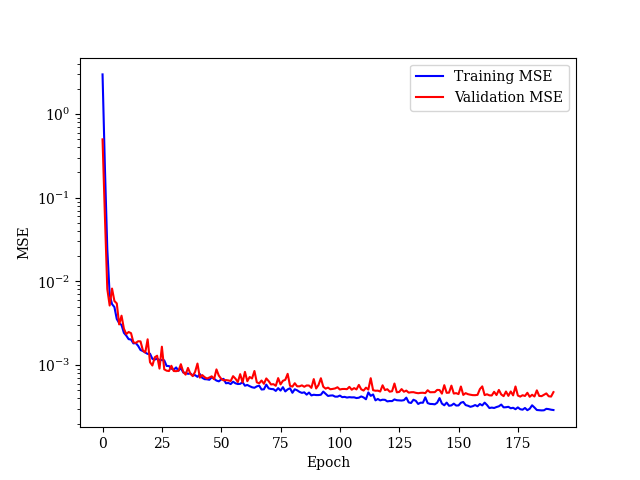}
	\caption{MSE value as a function of epoch during the training of Xception model for heat transfer prediction of single shape}\label{xception_mse}
\end{figure}

MSE values for each epoch of training and validation of optimized CNN model and Xception model are shown in Fig. \ref{cnn-loss2}, and Fig. \ref{xception_mse}, respectively. As it can be seen in these figures, Xception model has lower final MSE value. We provide more statistical analysis of the models fo single shape and multi-shape case in the next section. 



\subsubsection{Prediction performance}

Here, the performance of the surrogate models are analyzed using $R^2$ value also known as coefficient of determination which is computed by Eq. \ref{eq: r2}. 

\begin{equation}\label{eq: r2}
	R^2  =  1 - \frac{\Sigma_{i} {({{{{y}}}_i} -{{\hat{{y}}}_i})}^2}{\Sigma_{i} {({{{{y}}}_i} -\overline{{{{{y}}}}})}^2}
\end{equation}

where $\overline{{y}}$ is the baseline prediction which is considered to be the mean of ground truth values.

MAE is also used to evaluate the model accuracy according to Eq. \ref{eq: mae}
\begin{equation}\label{eq: mae}
	MAE = \frac{\Sigma_{i=1}^{N} ( {{{{y}}}_i} -{{\hat{{y}}}_i})}{N}
\end{equation}

\begin{figure}[h]
	\center
	
	\includegraphics[width=0.5\textwidth]{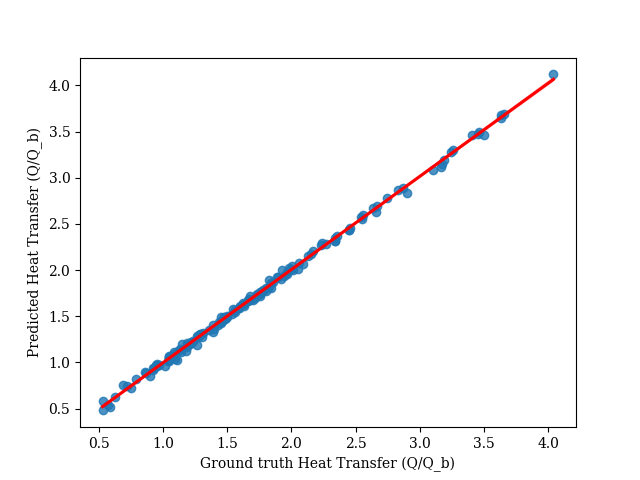}
	\caption{Predicted and ground truth heat transfer in presence of a single fin using optimized CNN model }\label{confidence_Q_cnn}
\end{figure}

\begin{figure}[h]
	\center
	
	\includegraphics[width=0.5\textwidth]{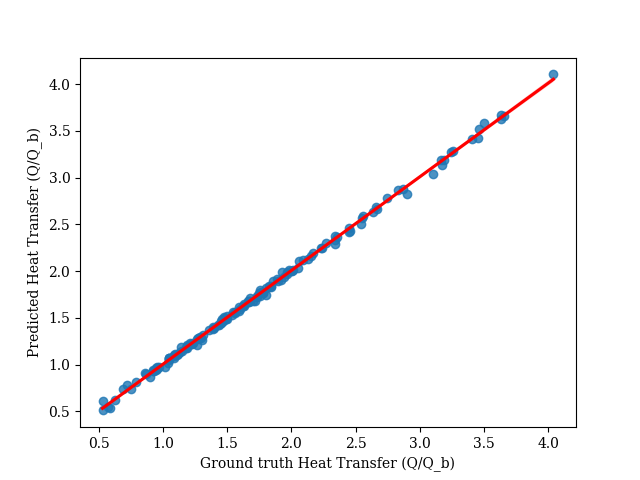}
	\caption{Predicted and ground truth heat transfer in presence of a single fin using Xception network }\label{confidence_Q_xception}
\end{figure}

In order to better visualize the estimated values, confidence intervals and residual plots are also used. Fig. \ref{confidence_Q_cnn} and Fig. \ref{confidence_Q_xception} show the predicted and ground truth values for heat transfer in presence of a single fin using optimized CNN model, and Xception model, respectively. On each plot, 150 data points of the same shapes in the test dataset are used to avoid clutter. 99 percent confidence intervals are also plotted in Fig. \ref{confidence_Q_cnn} and Fig. \ref{confidence_Q_xception} which are barely visible. This shows high accuracy in prediction. The trained model generalizes well beyond the training dataset as can be seen from the predicted values from the test dataset.

\begin{figure}[h]
	\center
	
	\includegraphics[width=0.5\textwidth]{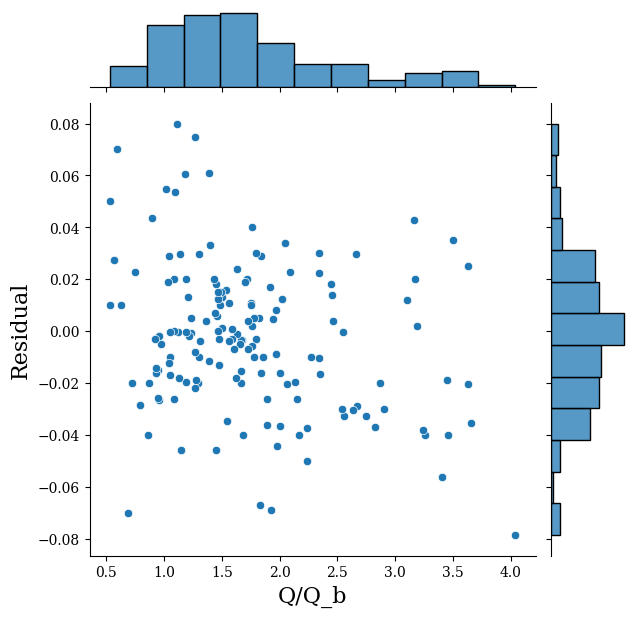}
	\caption{Residual plot for heat transfer estimation in presence of a single fin using optimized CNN model }\label{res_density_cnn}
\end{figure}

\begin{figure}[h]
	\center
	
	\includegraphics[width=0.5\textwidth]{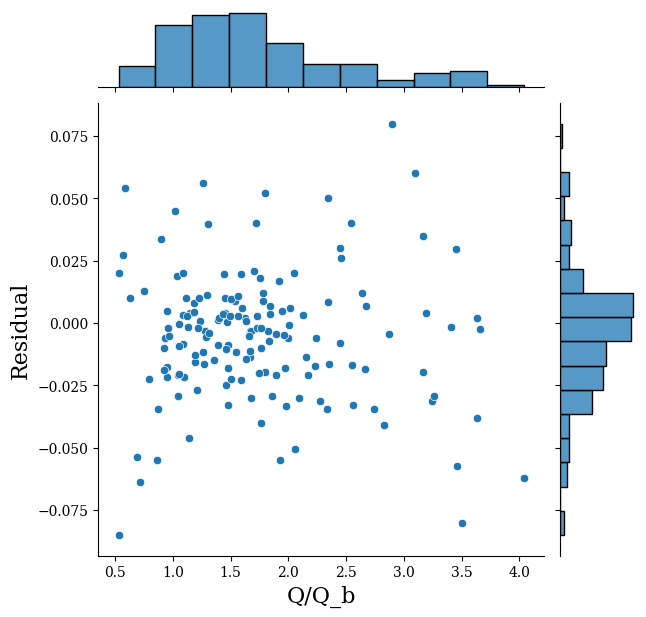}
	\caption{Residual plot for heat transfer estimation in presence of a single fin using Xception network}\label{res_density_xception}
\end{figure}

Fig. \ref{res_density_cnn} and Fig. \ref{res_density_xception} show the residual plot for heat transfer estimation models. Residual is defined as a difference between the predicted and actual values in the test dataset. Distribution of the residuals and data are also shown on the right hand side and top of the plot, respectively. One important observation from residual plot is that shapes with smaller heat transfer have higher deviation from the actual values. Higher prediction errors are associated with smaller heat transfer values which are pertinent to smaller shapes. This might be addressed by choosing the norm based on the shape area and normalizing the input data based on the shape area computed using Green's theorem \cite{yang1996fast}. However, area normalization was not performed since the accuracy of the models are within a great range. This normalization can be applied for higher complexity physics problems. 

Table \ref{stat_heat_single} shows the statistical summary for both heat transfer estimation models used for single shape. As it can be seen from the table, both models provide high accuracy prediction with Xception network having slightly higher performance. The coefficient of determination value is close to one and very hard to achieve in regression models.
\begin{table}[h]
	\centering
	\caption{Statistical results summary of the heat transfer surrogate models for a single morphable shape }\label{stat_heat_single}
	
	\begin{tabular}{ c|c|c }
		\hline
		\textbf{Model}  	&	\textbf{$R^2$} & MAE\\ 
		\hline
		
		CNN  & 0.997 & 0.022\\
		Xception & 0.998 & 0.019 \\

	\end{tabular}
	
\end{table}

\begin{figure}[h]
	\center
	
	\includegraphics[width=0.5\textwidth]{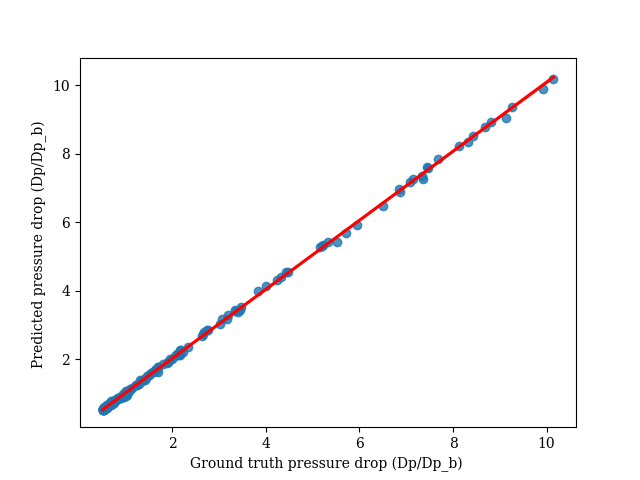}
	\caption{Predicted and ground truth pressure drop for a single shape using optimized CNN model (99 percent confidence interval is also plotted)  }\label{dp_dpb_cnn_confidence}
\end{figure}

\begin{figure}[h]
	\center
	
	\includegraphics[width=0.5\textwidth]{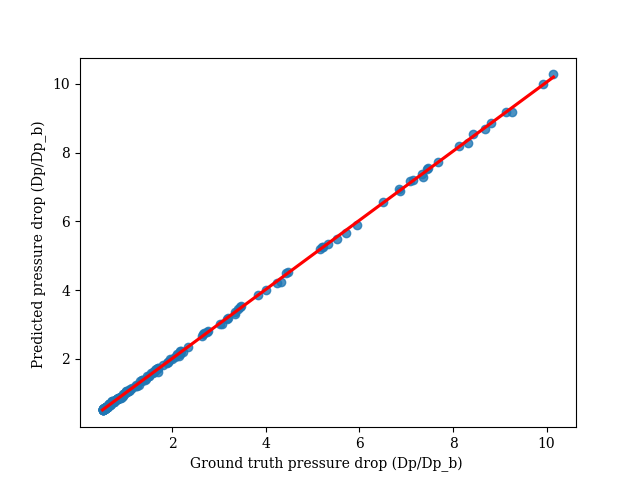}
	\caption{Predicted and ground truth pressure drop for a single shape using Xception network (99 percent confidence interval is also plotted) }\label{dp_dpb_xception_confidence}
\end{figure}

Fig \ref{dp_dpb_cnn_confidence} shows the predicted and ground truth values for pressure drop estimation for a single shape using optimized CNN for 200 data points. Fig. \ref{dp_dpb_xception_confidence} shows the estimated values for Xception network. It can be seen that predicted heat transfer values are closer to the regression line compared to pressure drop values for both models. This is because of the data distribution of the pressure values. Even though the same shapes are used for training the models, distribution of the CFD results  for heat transfer values are more compact than pressure drop values.

\begin{figure}[h]
	\center
	
	\includegraphics[width=0.5\textwidth]{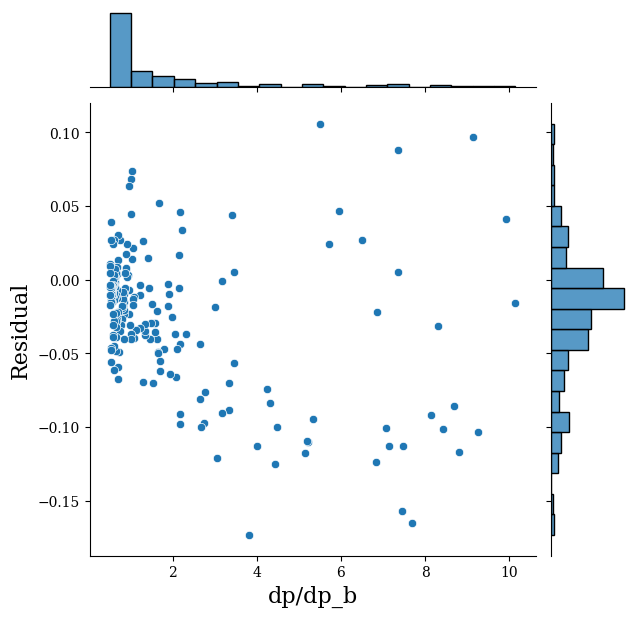}
	\caption{Residual plot for pressure drop estimation of a single fin using optimized CNN model }\label{res_density_cnn_p}
\end{figure}

\begin{figure}[h]
	\center
	
	\includegraphics[width=0.5\textwidth]{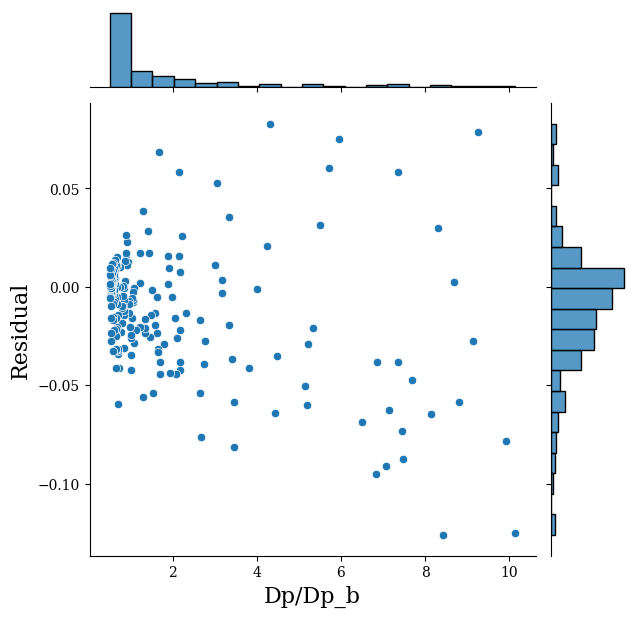}
	\caption{Residual plot for pressure drop estimation of a single fin using Xception network }\label{res_density_xception_p}
\end{figure}

Fig. \ref{res_density_cnn_p} and Fig. \ref{res_density_xception_p} show the residual for pressure drop using CNN and Xception network, respectively. Residual plot for optimized regular CNN shows very small bias in predicting pressure drop. The histogram is far from being right-skewed, but the peak distribution is not exactly at zero line. Small overestimation can be observed in the prediction which can be accumulated over the test set and cause larger MAE. Table \ref{stat_p_single} shows the statistical summary for the pressure drop prediction of a single shape. As it was mentioned, CNN model has a larger MAE. From the residual plot and MAE, it can be inferred that Xception works better for both pressure drop and heat transfer prediction.  

\begin{table}[h]
	\centering
	\caption{Statistical results summary of the pressure drop surrogate models for a single morphable shape }\label{stat_p_single}
	
	\begin{tabular}{ c|c|c }
		\hline
		\textbf{Model}  	&	\textbf{$R^2$} & MAE\\ 
		\hline
		
		CNN  & 0.999 & 0.0378\\
		Xception & 0.999 & 0.023 \\
	\end{tabular}
	
\end{table}

Table \ref{timetable} shows the average time required for the CNN models and CFD solver. Regular CNN provides faster computation compared to other methods. CFD solver occasionally fails because of the failure in mesh adaptation during the shape morphing in presence of sharp edges or acute angles. DL-based computation does not require mesh generation and provides more robust algorithm. High fidelity simulation has higher standard deviation in terms of computing time since variation of the shape causes change in number of meshes which leads to change in computation time. 
\begin{table}[h]
	\centering
	\caption{Computation time for different computing methods for a single fin }\label{timetable}
	
	\begin{tabular}{ c|c }
		\hline
		\textbf{Model}  	&	\textbf{Average prediction time} \\ 
		\hline
		
		CNN  & 0.42 seconds\\
		Xception & 2.03 seconds \\
		High fidelity simulation & 45 minutes\\

	\end{tabular}
	
\end{table}

\subsection{Multiple fins in domain}

\begin{figure}[h]
	
	\centering
	
	\includegraphics[width=.5\linewidth]{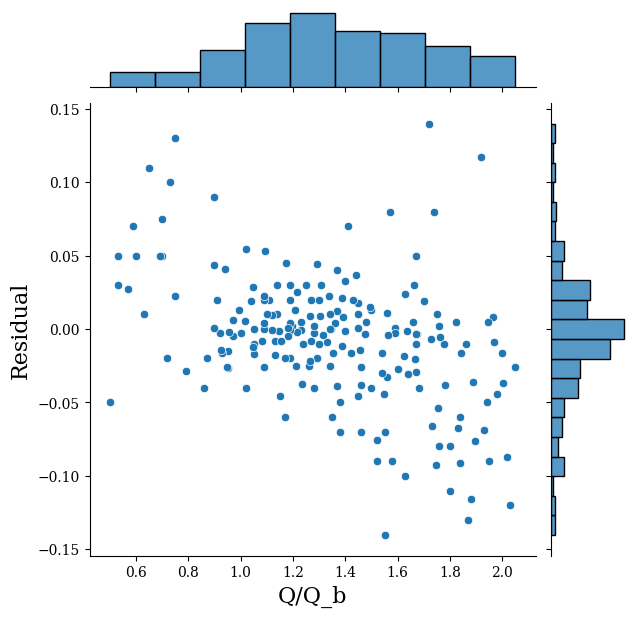}

	\caption{Residual plot of heat transfer prediction for multiple fin shapes}	\label{res_density_Q_marl}
\end{figure}

\begin{figure}[h]
	
	\centering
	
	\includegraphics[width=.5\linewidth]{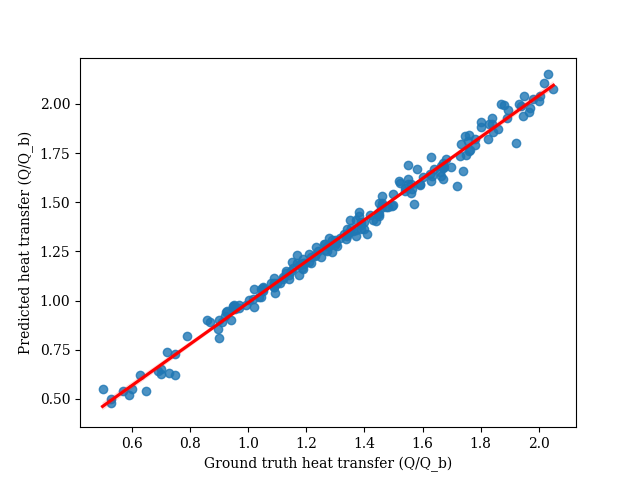}

	\caption{Predicted and ground truth heat transfer for multiple shapes (99 percent confidence interval is also plotted) }	\label{confidence_Q_marl}
\end{figure}

In this section, the performance of the surrogate models for multiple shapes across the domain are shown. Fig. \ref{res_density_Q_marl} shows the residual for heat transfer test dataset with the size of 200 images to avoid clutter. Comparing the surrogate model performance to the ones presented in previous section, it can be inferred that by increasing the number of shapes and consequently DOF, accuracy of the machine learning model decreases. Challenges to improve the ML model for higher accuracy remains open for ML community and competitions. Our takeaway from the error behavior is that increasing design freedom is associated with sacrificing accuracy in prediction. Higher DOF in design requires larger dataset for training the surrogate model. The model shows a reasonable performance with   $R^2$  values of 0.968 and MAE of 0.032. Residual plot shows that overall bias of the model can be considered very little to none. Normal distribution of the residual with a center at almost zero value shows the high accuracy of the model. The distribution of the data on top of the chart in Fig. \ref{res_density_Q_marl} shows that the center of the distribution is around the $Q/Q_b$ value of 1.3 which is associated with minimum residual as well. This provides high accuracy around this point. In general, the model overestimates the larger values of heat transfer and underestimates the smaller values. However, order of errors are small and can be considered as an accurate model. Fig \ref{confidence_Q_marl} shows predicted and ground truth values with 99 percent confidence interval. As can be seen, the model can be used for heat transfer prediction with high confidence.  

\begin{figure}[h]
	
	\centering
	
	\includegraphics[width=.5\linewidth]{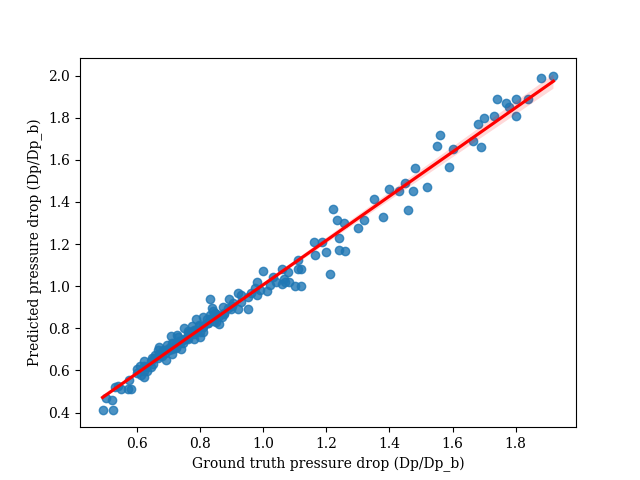}

	\caption{ Predicted and ground truth values of pressure drop with 99 percent confidence interval for multiple shape using Xception network}	\label{confidence_p_marl}
\end{figure}

Xception network is also used for pressure drop prediction of multiple fin shapes with minimum hyperparameter search and without any specific changes. Further details are provided in \ref{Appendix: xception_marl}.  Fig \ref{res_density_p_marl} shows the residual plot for Xception network trained for pressure drop prediction. Residual plot for pressure drop shows almost a normal distribution centered around zero value which indicates the model is fitted properly. Coefficient of determination is reported to be 0.98. MAE value of 0.033 is reported which is close to the heat transfer prediction value.    
\begin{figure}[h]
	
	\centering
	
	\includegraphics[width=.5\linewidth]{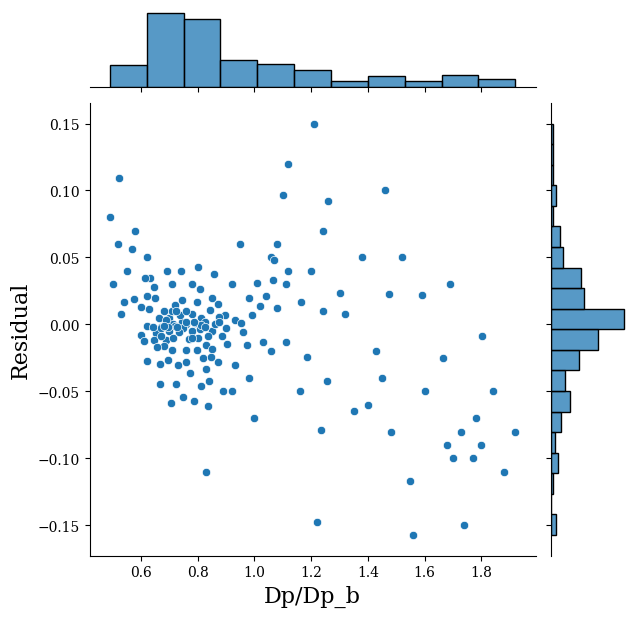}

	\caption{Residual plot of pressure drop values for multiple shape surrogate model}	\label{res_density_p_marl}
\end{figure}
Fig. \ref{confidence_p_marl} shows estimated values of pressure drop and actual values computed using high fidelity simulation. The red line shows the regression line with 99 percent confidence interval around the line. Less confidence around the higher values of pressure drop are seen. However, 99 percent confidence is of great value for surrogate modeling tasks.

\clearpage
\section{Conclusion and perspectives}

In this study, modern convolutional neural networks were utilized to predict heat transfer and pressure drop of the fins generated using composite Bezier curves without expensive CFD computation. An optimized CNN network using regular convolutions as well as an Xception model were deployed to predict heat transfer and pressure drop values of a single fin shape. Xception network showed higher accuracy in prediction compared to regular CNN. Xception network was used for the domain with multiple fin geometries. The accuracy of the surrogate models are reported and compared to the case with a single fin shape. Increase in design freedom of the thermal domain results in higher error in prediction. Larger dataset is also required to train the model for the domain with higher degrees of freedom. The errors in prediction remain within 3\% even for multiple fin shapes. These values are promising for surrogate modeling of morphable geometries.
In this paper, time-averaged CFD results were predicted directly from images of the fin geometries without using mesh representation.
This provides acceleration for heat transfer and pressure drop computation from several minutes to less than two seconds. The results presented in this study can be directly used for design optimization to accelerate the iteration process.

\par\noindent  The authors plan to use the presented surrogate models as the computational engine for multi-agent reinforcement learning framework for design automation of heat exchangers. This design automation is not possible without using a surrogate model.  

\section*{Declaration of competing interest}\noindent
The authors declare that they have no known competing financial interests or personal relationships that could
have appeared to influence the work reported in this paper.

\section*{Acknowledgment}\noindent
The authors thank Mojtaba Barzegari for his constructive feedback on FEniCS implementation. 

\section*{CRediT authorship contribution statement}
\noindent\textbf{Hadi Keramati}: Conceptualization, Methodology, Software, Validation, Writing and Editing\\
\textbf{Feridun Hamdullahpur}: Supervision, Reviewing and Editing\\
\clearpage

\appendix
\section{Hyperparameters for optimized regular CNN }\label{Appendix: cnnhyperparameters}
In table \ref{tablea2}, we provide hyperparameters of the optimized regular CNN to facilitate the reproducibility.  
\begin{table}[htb]
	\centering
	\caption{\label{tablea2} Hyperparameter values for optimized regular CNN}
	
	\begin{tabular}{ c|c }
		\hline
		\textbf{Hyperparameter}  	&	\textbf{Value} \\ 
		\hline
		
		Learning rate  & $1 \times 10^{-3}$\\
		Decay & $5 \times 10^{-3}$\\
		Number of Convolution layers & 11\\
		Number of MaxPooling layers & 5\\
		Number of FC layers & 2\\
		Activation function hidden layers&  ReLU\\
		Activation function output layer  & Linear\\
		Optimizer  & Adam\\		
		Batch size $ (N) $ & 128\\
		
	\end{tabular}
	
\end{table}

\newpage

\section{Optimized CNN architecture}\label{Appendix:c}
In this section, we show the architecture of the optimized  CNN used for the case with a single shape. The input image size is downscaled version of the original image by a factor of two with the size of 253 $\times$ 253 pixels. Activation functions for convolutional layers and the fully connected layer are ReLU and the output neuron is using linear function.
\begin{figure}[htb]
	\center
	
	\includegraphics[width=0.24\textwidth]{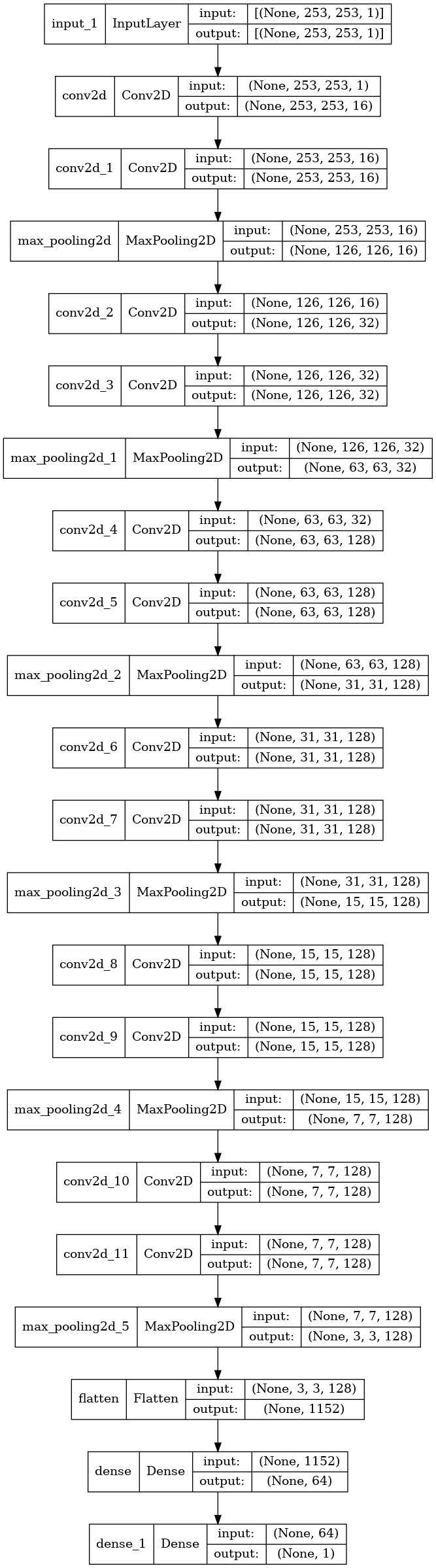}
	\caption{Optimized CNN architecture}\label{model_plot}
\end{figure} 
\newpage

\pagestyle{empty}

\section{Hyperparameters for Xception network used for single shape}\label{Appendix: xception}
In table \ref{tableb3}, some of the hyperparameters of the Xception model used for single fin shape are provided. Other hyperparameters are used the same as those in the original article \cite{chollet2017xception}.  
\begin{table}[htb]
	\centering
	\caption{\label{tableb3}Hyperparameters for Xception network used for single shape}
	
	\begin{tabular}{ c|c }
		\hline
		\textbf{Hyperparameter}  	&	\textbf{Value} \\ 
		\hline
		
		Learning rate  & $1 \times 10^{-3}$\\
		Optimizer  & SGD\\		
		Batch size $ (N) $ & 256\\
		
	\end{tabular}
	
\end{table}

\newpage

\section{Hyperparameters for Xception network used for multiple shapes}\label{Appendix: xception_marl}
In table \ref{tablebmarlhyp}, some of the hyperparameters of the Xception model used for multiple fin shapes are provided. No specific changes were made to the original architecture \cite{chollet2017xception}.  
\begin{table}[htb]
	\centering
	\caption{\label{tablebmarlhyp} Hyperparameters used for Xception network for multiple shapes}
	
	\begin{tabular}{ c|c }
		\hline
		\textbf{Hyperparameter}  	&	\textbf{Value} \\ 
		\hline
		
		Learning rate  & $0.0001$\\
		Optimizer  & Adam $(\beta_1 = 0.5, \beta_2 = .997)$\\		
		Batch size $ (N) $ & 256\\
		
	\end{tabular}
	
\end{table}

\clearpage

\bibliography{mybibfile}

\end{document}